# Stable In-hand Manipulation with Finger Specific Multi-agent Shadow Reward

Lingfeng Tao#, Jiucai Zhang^, and Xiaoli Zhang*, *Senior Member, IEEE*

*Abstract* — Deep Reinforcement Learning has shown its capability to solve the high degrees of freedom in control and the complex interaction with the object in the multi-finger dexterous in-hand manipulation tasks. Current DRL approaches prefer sparse rewards to dense rewards for the ease of training but lack behavior constraints during the manipulation process, leading to aggressive and unstable policies that are insufficient for safety-critical in-hand manipulation tasks. Dense rewards can regulate the policy to learn stable manipulation behaviors with continuous reward constraints but are hard to empirically define and slow to converge optimally. This work proposes the Finger-specific Multi-agent Shadow Reward (FMSR) method to determine the stable manipulation constraints in the form of dense reward based on the state-action occupancy measure, a general utility of DRL that is approximated during the learning process. Information Sharing (IS) across neighboring agents enables consensus training to accelerate the convergence. The methods are evaluated in two in-hand manipulation tasks on the Shadow Hand. The results show FMSR+IS converges faster in training, achieving a higher task success rate and better manipulation stability than conventional dense reward. The comparison indicates FMSR+IS achieves a comparable success rate even with the behavior constraint but much better manipulation stability than the policy trained with a sparse reward.

## I. INTRODUCTION

Dexterous in-hand manipulation is one of the essential functions for robots in human-robot interaction [1], intelligent manufacturing [2], telemanipulation [3], and assisted living [4], but it is also hard to solve due to the high degrees of freedom in control space and the complex interaction with the object. Deep Reinforcement Learning (DRL) [5] has shown its abilities in recent research [6-8] to solve dexterous in-hand manipulation tasks thanks to its learning capability, which enables the robot to find a control policy by interacting with the environment through exploration and exploitation.

The reward function is the guide to help the DRL policy learn how to solve the tasks. Researchers usually take time to design the reward function for DRL methods because it is one of the most critical factors affecting the training outcome. There are generally two forms of the reward function: one is sparse rewards, and the other is dense rewards [9]. When solving in-hand manipulation tasks with DRL approaches, the mainstream methods [10] prefer sparse reward functions, which give reward signals at the end of a period to evaluate the policy performance. The benefit of sparse reward is that with such a loose constraint on the policy behavior, training from the sparse reward is usually faster and easier to converge to a high success rate policy [10]. However, from our previous study [11], the sparse reward can also lead to aggressive and unstable manipulation behaviors such as finger cross, fingertip grasp, slippery edge, and less contact that may cause objects to fall and damage. Such unstable behaviors are insufficient in safety-critical tasks in real-world applications. For example, when manipulating fragile products (e.g., food and electronics), falling products will cause unrecoverable damage. Thus, it is necessary to help the robot to learn stable in-hand manipulation behaviors.

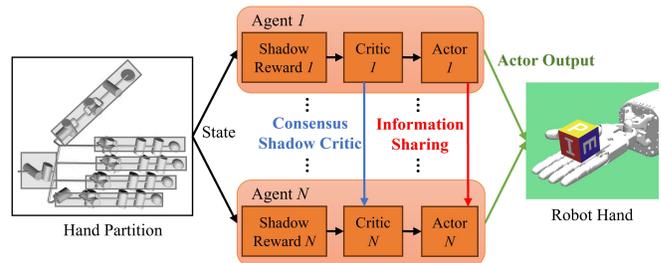

Fig. 1. An illustration of the multi-agent approach with FMSR and IS. The hand is partitioned to six agents. We assign a single agent to control one of the fingers and an additional agent to control the wrist. The critic of each agent has a global observation to the robot hand and all actions of other agents. The actor of each agent can observe the actions of its neighbors. Each agent has its specific shadow reward. The shadow gradient is calculated to update the critics and achieve consensus training through shadow critic. IS is applied across actors. The connection in this figure is designed for simplicity and ease of reading, which do not correspond to the actual settings.

Dense reward functions continuously evaluate the policy behavior and provide rich reward signals during the DRL training. Dense reward has been widely used in learning-based robot control problems, such as trajectory planning and object manipulation with robot arms [12][13], safety-aware navigation for mobile robots [14], and object grasping with a gripper [15]. These studies prove dense reward effectively guides the DRL policy to learn desired behaviors. However, recent literature [10] has pointed out dense reward is more challenging for in-hand manipulation tasks than sparse reward. Dense reward was used in [6] for generalizable real-world in-hand manipulation. The training requires 100 years of experience generated on a supercomputer. The reasons are twofold. First, compared to simple robot control tasks, it is hard to empirically define reward to guide the DRL policy not only to learn to complete the in-hand manipulation task but also to refine its manipulation behaviors. An inappropriately defined reward function can easily bias toward a specific policy that limits performance. Second, the training process of DRL is an optimization process that maximizes the reward function. The more complex the task and reward are, the

This material is based on work supported by the US NSF under grant 1652454 and 2114464.

#L. Tao is with Oklahoma State University, 563 Engineering North, Stillwater, OK, 74075, USA. (email: lingfeng.tao@okstate.edu)

*X. Zhang is with Colorado School of Mines, Intelligent Robotics and Systems Lab, 1500 Illinois St, Golden, CO 80401, USA. (e-mail: xlzhang@mines.edu).

^J. Zhang is with the GAC R&D Center Silicon Valley, Sunnyvale, CA 94085, USA. (e-mail: zhangjiucai@gmail.com).

harder it is to train a single optimal policy, as the solution is no longer pure convex, and no global optimal solution exists. How to enable dense reward in in-hand manipulation to help the policy to learn stable manipulation behavior is still an open problem.

This work proposes the Finger-specific Multi-agent Shadow Reward (FMSR) method (Fig. 1) to define the stable manipulation constraints based on the state-action occupancy measure, describing the probability distribution of state-action pairs an agent encounters when following its policy. As a learning property of DRL, the state-action occupancy measure can be treated as an equivalent of the behavior distribution, which is approximated during the training. The state-action occupancy measure is an appropriate candidate to construct reward objectives for behavior constraints as its convex distribution of the policy behavior [16] is convergent to the optimal solution [17]. FMSR adopts Multi-agent Shadow Reward [18] to define stability-aware behavior constraints for stable in-hand manipulation. The uniqueness of FMSR is that it extends the task objective from pure exploration to user-defined task objectives, which is finger cooperation to manipulate the object and maintain stability in our case. We embedded FMSR into the learning framework from our previous work Multi-agent Global-Observation Critic and Local-Observation Actor (MAGCLA) [11], modeling the in-hand manipulation as a multi-agent finger cooperation task and defining the observation relationship between neighbor fingers. In [11], a shared sparse reward was used. With FMSR, each agent is assigned a finger-specific shadow reward based on the task priorities of different fingers. Thus, instead of finding a globally optimal solution in a single-agent approach, the agents find the optimal local policies that can cooperatively complete the task. We then enable Information Sharing (IS) [18] across neighboring agents for consensus training and accelerating the training convergence. In summary, the contributions of this work are:
1) Developed FMSR based on the state-action occupancy measure to constrain the stable manipulation behavior.
2) Embedded FMSR to the MAGCLA multi-agent learning framework for specific reward design for each agent and enabling finger cooperation learning.
3) Designed IS across neighboring agents for consensus training and accelerating the training convergence.
4) Validated FMSR+IS on the Shadow dexterous hand in two in-hand manipulation tasks and compared their task performance and manipulation stability in the ablation study and the sparse reward with multiple objective stability-related evaluation metrics from literature.

## II. RELATED WORK

### A. Learning-based In-hand Manipulation
The rapid development of dexterous robotic hands has provided hardware foundations, such as the Shadow hand [6], an anthropomorphic hand with 24 degrees of freedom (DoFs), in which 20 joints are independently controllable. Tactile sensors like temperature [19], Hall effect [20], and electroactive polymeric [21] are developed to improve the fidelity of the robot's observation. With the readiness of robot hardware, researchers have been putting efforts into developing generalizable and adaptable methods for in-hand manipulation applications. DRL has demonstrated its capability to handle in-hand manipulation [6-8]. The OpenAI Gym [22] implements challenging in-hand manipulation tasks with the Shadow robot hand as a standard benchmark.

### B. Efforts to Improve Manipulation Stability
Recent research efforts to improve manipulation stability are applied in several ways. The first one is to design the reward function. [6] uses a dense reward for the task objective and a sparse reward that gives a huge penalty if the object dropped to improve the manipulation stability. Due to the difficulty in defining and learning, although it is desired, the dense reward that directly constrains the manipulation behaviors is rarely studied [6]. The other form to improve the manipulation stability is to increase the robustness of the control policy using methods like Domain Randomization [23]. The idea is to add instability (disturbance and noise) to the environment. There are other approaches to address instability in the real world. For example, when transferred to the real world, [6] slows down the control frequency to ensure that the robot can accurately follow the policy command and improve stability. In [24], human demonstrations are used to initialize the policy and are intentionally slowed down while collecting data. However, these approaches are trying to address the system instability rather than directly helping the policy to learn stable behaviors. Constrained behaviors learning, such as stable in-hand manipulation, are rarely studied.

## III. METHODOLOGY
This section introduces the representation of the multi-agent in-hand manipulation task in III. A. The FMSR development is explained in III. B. The IS method is shown in III. C.

### A. Multi-agent Modeling and Representation
The learning framework in this work is based on the MAGCLA algorithm developed in our previous work [11]. MAGCLA models the multi-agent in-hand manipulation as a Markov game [25], a multi-agent extension of the Markov Decision Process [26]. The Markov game contains $N$ agents, a set of action space $A_1, \ldots, A_N$, and a set of observations $O_1, \ldots, O_N$ that are assigned to each agent. Each agent follows a policy $\pi_{\theta_i}: O_i \times A_i \mapsto [0,1]$. A state $S$ is defined to describe the Markov game. The execution of all agents' actions produces the transition to the next state by following the state transition function $\Gamma: S \times A_1 \times \ldots \times A_N \mapsto S'$. For multi-agent in-hand manipulation, the action space $A_i$ of agent $i$ is based on the hand partitions (Fig. 1). In this work, we assign each agent to control one of the fingers and an additional agent to control the wrist. The observable state is denoted as $x$, including the positions and velocities of the joints and the Cartesian position and rotation of the object represented by a quaternion as its linear and angular velocities.

### B. Finger-specific Multi-agent Shadow Reward
For a multi-agent setup, we define the local state-action occupancy measure $\lambda_i^\pi(s_i, a_i)$ for agent $i$ as:

$$\lambda_i^\pi(s_i, a_i) = \sum_{t=0}^{T} \gamma^t \, \mathbb{p}(s_i^t = s_i, a_i^t = a_i | \pi_i, s_i^0 \sim \xi, g \sim G) \quad (1)$$

where $i \in [1,2,\ldots,N]$ is the index for $N$ agents, $\pi_i$ is the policy for agent $i$, $s_i^0$ is the initial state sampled from distribution $\xi$, $g$ is the goal sampled from distribution $G$. $\gamma$ is a discount factor, $t$ is the time step, and $T$ is the maximum time step. $\mathbb{p}$ is the probability distribution that the agent choose action $a_i$ at state $s_i$ in time $t$ when start with policy $\pi_i$, initial state $s_i^0$, goal $g$. A shadow reward function $F_i$ for agent

$i$ is defined based on the local state-action occupancy as $F_i(\lambda_i^\pi(s_i, a_i))$. The detailed stability-related shadow reward function will be explained in section IV for easier organization. A shared task-related reward function is designed for each agent based on the state and action $r_i : \times A_i \mapsto \mathbb{R}$, to help the agents to understand the task objective. The agent $i$ should maximize its expected total reward:

$$R_i = \sum_{t=0}^{T}\left(\gamma^t r_i^t + \alpha F_i(\lambda_i^\pi(s_i, a_i))\right) \quad (2)$$

where $\alpha$ is a scaler to weigh the shadow reward. FMSR adopts the deterministic policy gradient method [36] for continuous action space. For each agent $i$, we train a continuous actor $\mu_{\theta_i}$, where $\theta_i$ is the network parameters to maximize the objective:

$$J(\mu_{\theta_i}) = \mathbb{E}_{s \sim p^\pi}[R_i] \quad (3)$$

where $p^\pi$ is the state distribution. We can write the Q function to predict the potential state-action value of agent $i$ as:

$$Q_i^\mu(s_i, a_i) =$$
$$\mathbb{E}\left[\sum_{t=0}^{T} \gamma^t r_i^t + \varepsilon F_i(\lambda_i^\pi(s_i^t, a_i^t))|\pi_i, s_i^0 \sim \xi, g \sim G\right] \quad (4)$$

The gradient of the actor can be calculated as:

$$\nabla_{\theta_i} J(\mu_{\theta_i}) = \mathbb{E}_{x,a \sim D}[\nabla_{\theta_i}\mu_{\theta_i}(a_i)\nabla_{a_i}Q_i^\mu(s_i, a_i)] \quad (5)$$

where $D$ is the replay buffer which contains a transition tuple $(x, x', a_1, \ldots, a_6, r_1, \ldots, r_6)$. With the chain rule, the differentiation of shadow reward $F_i(\lambda_i^\pi)$ with respect to $\theta_i$ is:

$$\nabla_{\theta_i} F_i(\lambda^{\pi\theta_i}) =$$
$$\mathbb{E}\left[\sum_{t=0}^{T} \gamma^t Q_i^\mu(s_i^t, a_i^t) \cdot \nabla_{\theta_i} \log \pi_{\theta_i}(a_i^t|s_i^t)|\pi_{\theta_i}, s_i^0 \sim \xi, g \sim G\right] \quad (6)$$

The critic is updated by minimizing the loss function:

$$\mathcal{L}(\theta_i) = \mathbb{E}_{x,a,r,x'}\left[Q_i^\mu(s_i, a_i) - y\right]^2$$
$$\text{where } y = r_i + \varepsilon F_i(\lambda_i^\pi(s_i, a_i)) + \gamma Q_i^{\mu'}(s'_i, a'_i) \quad (7)$$

$\mu'$ is the target with delayed parameters $\theta'$ for stable updating.

### C. Information Sharing

In a multi-agent setup, because the in-hand manipulation task requires seamless cooperation during the manipulation process, all policies must follow the same pace to increase their performance, which means consensus training for all agents. To achieve this, we first use our experience synchronization method [11], which synchronizes the replay buffer for all agents and trains the policies with the same experience period when updating networks. Then, we further enable the IS method across neighbor agents so that the neighbor agents can directly share gradient information for policy updates. In literature, IS is employed to diffuse information between agents across time while optimizing their local policy [27, 28]. This scheme originates in flocking [29] and gossip protocols [30, 31], and may be interpreted as an approximate enforcement strategy for equality constraints of the network parameters held by distinct agents [18].

In this work, a right-side Shadow hand is used in the experiment. The gradient information is shared in clockwise order. The thumb and little finger are considered neighbors to ensure information effectively propagates across agents in a loop. It should be noted that more information sharing formation is possible, which will be studied in future research. All agents perform a simple weighted averaging step using mixing matrix $M$, a symmetric doubly stochastic matrix that

---

**Algorithm 1** FMSR and IS

**Initialize** critic $Q_i^\mu$, actor $\mu_{\theta_i}$ for agent $i = 1$ to $N$, replay buffer $D$, and random noise $\mathcal{N}$, Mixing matrix $M$, max-episode-length $T$.

1: **Start** episode **do**
2:     Initialize state $x$
3:     **for** $t = 1$ to $T$ **do**
4:         Select action $a_i = \mu_{\theta_i}(s_i, a_i) + \mathcal{N}$ for agent $i = 1,2,\ldots N$,
5:         Save transition $(x, a_1, \ldots, a_N, r, F_i(\lambda_i^\pi(s_i, a_i)), x')$ to $D$
6:         Set $x \leftarrow x'$
7:         **for** agent $i = 1$ to $N$ **do**
8:             Estimate occupancy measure $\lambda_i^\pi(s_i, a_i)$
9:             Update shadow reward $F_i(\lambda_i^\pi(s_i, a_i))$
10:            Update critic $Q_i^\mu$
11:            Information sharing $\theta'^{k+1}_i = \sum_{\{j:(j,i)\in\varepsilon\}} M(j,i) \cdot \theta_i^{k+1}$
12:            Update actor $\mu_{\theta_i}$
13:         **end for**
14:         Update target network parameters $\theta_i' \leftarrow \tau\theta_i + (1-\tau)\theta_i'$
15:     **end for**
16: **end** episode

---

respects the edge connectivity $\varepsilon$ of the graph based on the finger relation. When agents execute information per step $k$, the shared gradient is computed as:

$$\theta'^{k+1}_i = \sum_{\{j:(j,i)\in\varepsilon\}} M(j,i) \cdot \theta_i^{k+1} \quad (8)$$

After IS, each agent's target policy $\theta_i'$ are updated at the end of every epoch as:

$$\theta_i' \leftarrow \tau\theta_i + (1-\tau)\theta_i' \quad (9)$$

where $\tau$ is the learning rate, the overall proposed FMSR and IS are summarized in Algorithm 1, which proceeds in four stages: (1) approximate occupancy measure to obtain the shadow reward; (2) critic updates; (3) information sharing; and (4) actor updates.

## IV. EXPERIMENTS

### A. Task Design

The FMSR and IS will be evaluated in a simulated environment for easy training and testing. Specifically, we use the Shadow hand environments with 92 tactile sensors from the OpenAI Gym platform, which runs on the MuJoCo [33] physics simulator. We partition the Shadow hand into 6 agents: 5 agents control each finger, respectively, and the additional agent controls the wrist. Based on the hand partition, the 6 DRL agents are wrist (2 DoFs), thumb (5 DoFs), index (3 DoFs), middle (3 DoFs), ring (3 DoFs), and little (4 DoFs). The IS method is applied across all fingers. Two in-hand manipulation environments are designed to evaluate the generalizability of our methods in different tasks:

1) *Block manipulation*. A block is placed on the Shadow hand with a random initial pose (Fig. 2). The task is manipulating the block around the Z-axis to achieve the target pose.
2) *Egg manipulation*. The task is similar to block manipulation, but an egg-shaped object is used.

During training, a goal is considered achieved if the difference in the rotation is less than 0.1 rad. During testing, the criterion changes to 0.4 rad. Such a setting helps the policy achieve higher performance in training. The agents are running at a time step of 0.04s. The policies are trained with the Message Passing Interface (MPI) [32], a parallel training tool that can run multiple simulation threads to accelerate the training process. The PC hardware for training includes an Intel 12900K, an Nvidia RTX3080ti, and 64 GB of RAM. Most hyperparameters are from [10], but with changes to the number of MPI workers to 4, total epoch to 400, cycles per epoch to 25, and batches per cycle to 25 for less training time.

B. Shadow Reward Design

The finger-specific reward is constructed based on multiple reward components. The task objective reward component is:
$$R_1 = -|z_t - z_g| \quad (10)$$
where $z_t$ is the object's rotational position, and $z_g$ is the goal position. $R_1$ provides a penalty signal based on the distance between the object's current and goal positions.

Two shadow reward components are designed to help the policies learn stable manipulation behaviors. The first shadow reward component constrains the safe region by limiting the object to stay in the center area of the palm, which is:
$$R_2 = F_1(\lambda^\pi) = -\log(\lambda^\pi(\bar{S}) + 0.1), \bar{S} \in |\mu - c|_2 < \rho \quad (11)$$
where $\bar{S}$ defines the safety area to satisfy the condition that the distance between the object's center position $\mu$ and the palm center $c$ is less than threshold $\rho$. The coefficient 0.1 is to avoid the safety region dominating the reward signal.

The second shadow reward component constrains the number of contact points based on the rationale that the more contact points with the object, the stabler the manipulation is. We construct it as a shadow reward component, which does not directly constrain the number of contacts but provides a reward or penalty based on state-action pairs that generate more contact points. This shadow reward component is:
$$R_3 = F_2(\lambda^\pi) = -\log\left(\sum_{s,a} \lambda^\pi(s,a)_g \varphi_j(s,a)_g + 0.1\right) \quad (12)$$
where $\varphi_j(s,a)_g = 1$ if the number of contact points is greater than threshold $\tau$. In this work, we set $\tau = 10$, which means 10 tactile sensors among the total 92 sensors that should be activated during the manipulation.

With all the reward components, we can specify the reward function for every agent. We first define the reward function for the wrist:
$$\mathbb{R}_1 = \sum (R_1, R_2, R_3) \quad (13)$$
which means the wrist will learn all three objectives as it controls the palm movement and affects the whole hand's dynamics. The thumb and little fingers share the reward:
$$\mathbb{R}_2 = \sum (R_1, R_2) \quad (14)$$
Based on previous studies [11], the thumb and little fingers contribute more than other fingers to manipulating rotational tasks with active contacts. Thus, we expect these two fingers to focus on learning the task objectives and also keep the object within the safe area. The index, middle, and ring fingers tend to contact the object to avoid the object falling passively. Thus, we define their reward as:
$$\mathbb{R}_3 = \sum (R_1, R_3) \quad (15)$$

C. Evaluation Metrics

The following configurations were implemented for the ablation study and compared with baselines: 1) Dense+FMSR+IS, 2) Dense+FMSR, 3) Dense, and 4) Sparse. These experimental setups allow us to compare and test our two separate methods, FMSR and IS. Comparing 1) and 2) evaluate the improvements of the IS. Likewise, we compare 2) and 3). We then compare the improvements FMSR+IS has to the dense reward with 1) and 3). Lastly, we compare our methods with the sparse reward with 1) and 4).

During the training process, the validating set contains unlimited trials with randomly generated target positions within the range of $(-\pi, \pi)\,rad$. The task success rate is recorded at the end of each epoch for a direct and unbiased comparison. The success rate is the percentage of successful cases in validation over 50 trials that are randomly generated initial and target poses. Each configuration is trained 3 times, each time with 400 epochs, to obtain the statistical results. A testing set is also generated, containing 100 trials with random targets and initial poses unseen in training and validation. The testing set is reused in all evaluations for a reproducible comparison. To test the manipulation stability, we use the following objective evaluation metrics:

**The success rate in the testing set.** We expect performance sacrifice compared to training without shadow reward. However, it is worth analyzing the performance difference for future studies to improve manipulation stability while maintaining task performance.

**The number of contact points.** This measure can be directly read from the tactile sensors, which is practical for future evaluation on the physical robot hand. Two measures will be collected: one is the average number of contact points for a single episode over the testing set, which helps quantify the overall manipulation stability; the other is the temporal number of contact points, which helps to analyze the manipulation behavior in the time domain.

**Measures of dynamic manipulation stability.** In-hand manipulation is a dynamic control process in the time domain, making the manipulation stability vary when the robot interacts with the object. Evaluating the manipulation stability in the time domain from the initial to the target state is necessary. Thus, we adopt the following measures from the literature [34] to objectively assess the policy behavior regarding how the hand grasps the object. The first two measures are the minimum singular value $Q_{MSV}$ and the derivative $Q'_{MSV}$. $Q_{MSV}$ is defined as:
$$Q_{MSV} = \sigma_{min}(GS) \quad (16)$$
where $GS$ is a grasp matrix defined in [34] to describe the relationship between forces at the contact points, the total wrench applied on the object, and the relation between velocities at the contact points and the twist. The smallest singular value of the grasp matrix $GS$ is a quality measure that indicates how far the grasp configuration is from falling into a singular configuration [35]. A large $Q_{MSV}$ leads to a better grasp. The derivative $Q'_{MSV}$ indicates how fast the grasp configuration is changing. A small $Q'_{MSV}$ is preferred. The third and fourth measures are the volume of the wrench ellipsoid $Q_{VEW}$ and the derivative $Q'_{VEW}$. $Q_{VEW}$ is defined as:
$$Q_{VEW} = \sqrt{\det(GG^T)} = \sigma_1 \sigma_2 \ldots \sigma_d \quad (17)$$
where $\sigma_1 \sigma_2 \ldots \sigma_d$ denoting the singular values of the grasp matrix $G$. This measure indicates the global contribution of all contact forces, which considers all the singular values with the same weight and must be maximized to obtain the optimum grasp. Its derivative $Q'_{VEW}$ also indicates how fast the grasp configuration is changing in the wrench space. A small $Q'_{VEW}$ is preferred. The last measure is the average distance between contacts to the object centroid $Q_{DCC}$:
$$Q_{DCC} = \frac{1}{n}\sum_{i=1}^{n} dist(c_i, c_m) \quad (18)$$
where $dist(c_i, c_m)$ calculates the distance between contact points $c_i$ and the object mass center $c_m$. $Q_{DCC}$ indicates the compactness of power grasp. A small $Q_{DCC}$ means the hand is grasping the object firmly with the minimized effect of inertial and gravitational forces, which is considered stable.

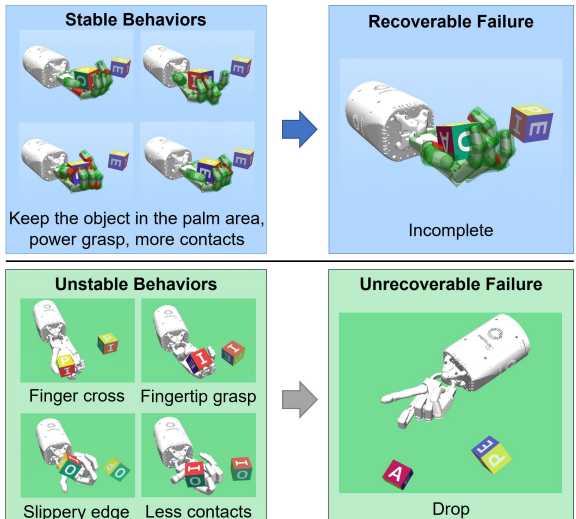

Fig. 2. (a) stable behaviors experience more of the failure type: **incomplete**, when the hand stops manipulating the object but still does not reach the target. (b) unstable behaviors experience more of the failure type: **drop**, where the object falls off the hand during manipulation.

**Failure type:** The failure type of the policy learned by the dense and sparse reward will be compared. In our tests, success means the policy completes the task in a defined number of steps for DRL training and testing. The failure means the policy fails to achieve the task objectives in a defined number of steps. In our case, task success is the robot hand rotates the object from the initial to the target position. Likewise, the task failure is when the robot fails to do so. In addition, we observed that the failure cases of dense and sparse policies are significantly different. To demonstrate the difference, we categorize the failure into two types. One is called **Incomplete**, and the other is called **Drop** (shown in Fig. 2). In the incomplete type, the hand stops to manipulate the object but does not reach the target. Because the object stays in hand, it will not be damaged, and it is possible to reset the hand configuration in real-world scenarios to redo the task. On the contrary, the object falls off the hand for the drop type, which is hard to recover. The damage will be expensive in real-world situations like manipulating fragile and high-value objects (e.g., jewelry, electronics). Thus, the Incomplete failure type is less severe than the Drop failure type.

## V. RESULTS AND DISCUSSION

### A. Training Process

The results of the training process are shown in Fig 3. We only show the training curves for dense reward functions for a fair comparison. The comparison between dense and sparse rewards can be found in [10]. Overall, the rotating block task is more difficult than the rotating egg task. The block's plane surface and sharp edges easily cause slippery movement and object falling. In contrast, the egg object has a symmetric shape and continuously curved surface, making the grasp stabler and easier to manipulate.

The multiple training processes in both tasks show consistent learning curves for all configurations. With a strict success criterion, Dense+FMSR+IS (blue) has the highest learning speed and converged success rate compared to other configurations. The baseline Dense (yellow) has the lowest learning speed and converged success rate. Dense+FMSR (red)'s learning speed and success rate are higher than the baseline Dense (yellow). The baseline methods Dense+FMSR (red) and Dense (yellow) achieved lower performance in the egg rotation task. A potential reason is that the egg used in the experiment is smaller than the block, making it harder to maintain a stable grasp during manipulation, increasing the difficulty of learning the stable manipulation behavior.

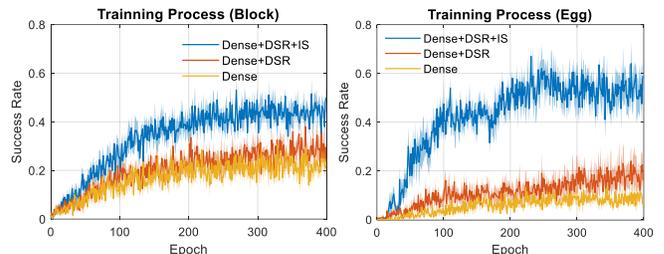

Fig. 3. The training process in block and egg rotating tasks. Dense+FMSR+IS (blue) has the highest learning speed and converged success rate compared to other configurations. The baseline Dense (yellow) has the lowest learning speed and converged success rate. Dense+FMSR (red)'s learning speed and success rate are slightly higher than the baseline Dense (yellow).

Comparing Dense+FMSR+IS (blue) and Dense+FMSR (red) proves that sharing the gradient information across neighbors can significantly benefit the training. Comparing Dense+FMSR (red) and Dense (yellow) shows that the shadow reward constraints have no negative effects on learning efficiency but slightly increase it, proving the shadow reward is an appropriate behavior constraint.

### B. Statistical Results

Table I shows the results of task performance and manipulation stability measures. The highest success rate among the trained policies for each configuration is logged.

**The success rate in the testing set.** Dense+FMSR+IS achieved the highest success rate in both tasks among all the dense rewards. It is slightly lower than the sparse reward, which matches our expectations due to the constrained search space. However, the later behavior analysis will reveal the necessity of performance sacrifice for stable manipulation.

**The number of contact points.** Table I shows the total number of contact points. Dense+FMSR+IS achieved around 20% higher number of contacts than other baseline methods, which means higher manipulation stability. The temporal contact is shown in the next subsection.

**Measures of dynamic manipulation stability.** On average, the Dense+FMSR+IS achieved the largest minimum singular value $Q_{MSV}$, the largest volume of the wrench ellipsoid $Q_{VEW}$, also proving the highest manipulation stability compared to other methods. For the first-order derivative $Q'_{MSV}$ and $Q'_{VEW}$, Dense+FMSR+IS has a lower rate of change except the $Q'_{MSV}$ in the egg rotation task. A possible explanation is that the egg task is to rotate around the Z axis, and the round bottom of the egg is naturally unstable. The robot needs extra effort to keep the egg in the vertical position. This behavior is reflected in the higher changing rate in $Q_{MSV}$. However, it does not affect the manipulation stability much since the $Q_{MSV}$ is much higher than other methods. Lastly, Dense+FMSR+IS achieved the lowest $Q_{DCC}$, showing signs

TABLE I. Task Performance and Manipulation Stability Measures over 100 Trials (The best performance is bolded)

| | | Success Rate | Total Contact | $\overline{Q_{MSV}}$ | $\overline{Q_{VEW}}$ | $\overline{Q'_{MSV}}$ | $\overline{Q'_{VEW}}$ | $\overline{Q_{DCC}} * e^{-3}(m)$ |
|---|---|---|---|---|---|---|---|---|
| Block | Dense+FMSR+IS | 86% | **7614±331** | 919±145 | 2644±223 | 10±60 | 61±173 | 6.9±2.4 |
| | Dense+FMSR | 67% | 5787±251 | 667±245 | 1558±247 | 58±122 | 91±171 | 15.6±3.6 |
| | Dense | 61% | 5359±263 | 434±85 | 1276±275 | 36±70 | 128±269 | 17.6±5.3 |
| | Sparse | **92%** | 5621±451 | 441±178 | 1685±436 | 48±82 | 86±131 | 19.3±9.2 |
| Egg | Dense+FMSR+IS | 91% | 6595±374 | **1072±173** | 2175±251 | 55±152 | 125±276 | **3.8±2.5** |
| | Dense+FMSR | 43% | 5846±269 | 912±151 | 1714±304 | 38±101 | 129±212 | 6.5±3.3 |
| | Dense | 35% | 5624±203 | 707±80 | 1266±341 | **23±68** | 191±245 | 8.1±1.9 |
| | Sparse | **94%** | 6019±496 | 764±213 | 1584±396 | 67±94 | 168±218 | 5.9±4.1 |

TABLE II. Compare Failure Types of Dense and Sparse Reward over 500 Trials

| | | # of F~ | Incom^ | Incom% | Drop | Drop% |
|---|---|---|---|---|---|---|
| Block | Ours | 61 | **60** | **98.4%** | 1 | **1.6%** |
| | Sparse | **42** | 26 | 61.9% | 16 | 38.1% |
| Egg | Ours | 53 | **53** | **100%** | **0** | **0** |
| | Sparse | **37** | 24 | 64.9% | 13 | 35.1% |

Ours is Dense+FMSR+IS. F~ is failure, Incom^ is incomplete. The best performance is bolded.

of stable power grasp. A better visualized comparison of the stable manipulation can be found in the supplemental video.

*C. Manipulation Stability Analysis*

We analyze the manipulation stability of all the trained configurations for block and egg rotation tasks in the time domain. All the plots are generated in the same testing case (shown in Fig. 4). More videos can be found in the supplemental materials.

Dense+FMSR+IS keeps consistently high $Q_{VEW}$ for both block and egg rotation tasks. The block test took less time because it's easier to stabilize the block at the target position. For $Q'_{VEW}$, Dense+FMSR+IS and Dense+FMSR need less adjustment during the manipulation. Due to unstable manipulation, the Dense policy applied large impulses to change the grasp configuration. Dense+FMSR+IS has a constant high $Q_{MSV}$ in block and egg tests, but the egg test shows a drop in $Q_{MSV}$ which corresponds to the large variance in $Q'_{MSV}$. This is because the hand needs to readjust the grasping configuration from dynamic manipulation to static grasping as the task is nearly completed, making the stability drop during the adjustment. For the temporal number of contact points, Dense+FMSR+IS can keep more than 10 contact points during manipulation for tests most of the time. For the average distance $Q_{DCC}$, Dense+FMSR+IS outperformed the baseline methods with a much smaller $Q_{DCC}$ when achieved the target position.

*D. Compare Failure Type of Dense and Sparse Reward*

In Table I, the task performance of the sparse reward policy is higher than Dense+FMSR+IS, although it has much higher manipulation stability in the evaluation metrics. In the authors' opinion, we cannot choose one approach over others according to the task performance only. When choosing dense or sparse rewards, multiple criteria must be considered for real-world applications. In this section, we look deeply into the manipulation behaviors of the policies trained with dense and sparse rewards.

To statistically compare the failure type difference, we test the policies trained with Dense+FMSR+IS and Sparse with 500 trials. The results are shown in Table II. In both tasks, Dense+FMSR+IS has slightly more failures than Sparse. However, considering the percentage, Dense+FMSR+IS only has a 1.6% and 0% drop rate in block and egg tasks. Considering the number, Dense+FMSR+IS dropped the object once over 1000 trials, and the Sparse policy dropped 29 times. The results further prove the importance of manipulation stability. Our suggestion is that when task success is the priority, training the policy with sparse reward is more beneficial. When regulating manipulation behavior is needed, training the policy with the proposed shadow reward approach provides stabler motions.

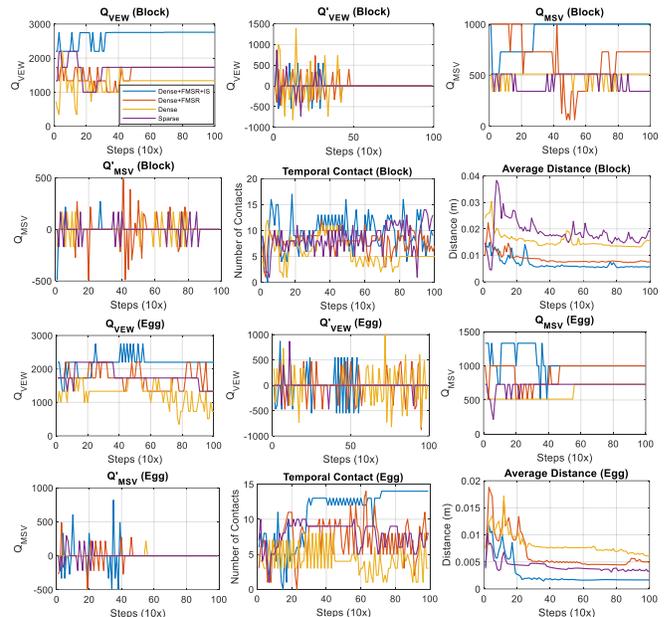

Fig. 4. The plots of the dynamic manipulation stability measures, which are generated in the same test case with data collected in every 10 steps.

## VI. Conclusion

This work proposed the Finger Specific Multi-agent Shadow Reward method to design the stability-related behavior constraint based on the state-action occupancy measure, avoiding empirically defining the reward function and biasing the training. Information Sharing is enabled across the neighbor fingers for consensus training and efficient convergence. The methods are evaluated with objective dynamic manipulation stability measures from the literature. We further compare the failure types of policies trained with dense and sparse rewards. The proposed methods can significantly improve manipulation stability and reduce the number of object drops, which can benefit real-world applications. Future research will focus on improving the task performance of dense reward approaches and studying different behavior constraints designed in shadow reward that can be applied to robotics applications.